# A novel illumination condition varied image dataset-Food Vision Dataset(FVD) for fair and reliable consumer acceptability predictions from food

Swarna Sethu
ssethu@uark.edu

Department of Biological and Agricultural Engineering
University of Arkansas Fayetteville, 72701, USA.

Dongyi Wang *
dongyiw@uark.edu

Department of Food Science &
Department of Biological and Agricultural Engineering
University of Arkansas Fayetteville, 72701, USA.
* corresponding author

Recent advances in artificial intelligence promote a wide range of computer vision applications in many different domains. Digital cameras, acting as human eyes, can perceive fundamental object properties, such as shapes and colors, and can be further used for conducting high-level tasks, such as image classification, and object detections. Human perceptions have been widely recognized as the ground truth for training and evaluating computer vision models. However, in some cases, humans can be deceived by what they have seen. Well-functioned human vision relies on stable external lighting while unnatural illumination would influence human perception of essential characteristics of goods. To evaluate the illumination effects on human and computer perceptions, the group presents a novel dataset, the Food Vision Dataset (FVD), to create an evaluation benchmark to quantify illumination effects, and to push forward developments of illumination estimation methods for fair and reliable consumer acceptability prediction from food appearances. FVD consists of 675 images captured under 3 different power and 5 different temperature settings every alternate day for five such days.

# Introduction

The design of cameras aims to mimic the functioning of human eyes for recording real-world scenarios in a visual image. Recent advances in deep learning have brought intelligence into the digital image understanding. In many specific tasks, intelligent algorithms have been approaching or exceeding human level performance (LeCun et al., 2015). Although human perceptions have been widely recognized as the ground truths for training and evaluating different deep learning-based image understanding models, in some cases, humans can be deceived by what they have seen, because human vision relies on stable external lighting to perceive fundamental object properties such as shapes and colors (Murray and Adams, 2019). Unnatural illumination would influence human perceptions of essential characteristics of goods (Wang et al., 2015). Under these conditions, 'well-trained' artificial intelligence models can only 'naively' repeat human 'mistakes.' An everyday example of the illumination effects is, in restaurant and retail stores, when food products are placed under different lighting conditions, humans will feel differently in response to the products, which can further affect consumers' consumption decisions (Hasenbeck et al., 2014). Therefore, this study aims to build a novel dataset, the Food Vision Dataset (FVD), to create an evaluation benchmark to quantify illumination effects on human and computer perceptions.

# 1.Dataset Description

## 1.1 Sample preparation

In the experiment, lettuce, a common fresh food product, is regarded as the experimental food sample. Lettuce is one of the most valuable fresh vegetables and is in the top ten most valuable crops in the U.S., with an annual farm-gate value of over $2.3 billion ("Andrew-LaVigne-House-Ag-testimony-7-12-17.pdf," n.d.). The romaine lettuce samples are randomly purchased from local grocery stores, and prepared and sliced by research assistants at the University of Arkansas. Romaine lettuces are widely used in fast food supply chains and are linked to many E. coli O157:H7 outbreaks. In the experiment, a total of 9 lettuce samples are purchased at different dates in different grocery stores around the campus. For each sample, fifty grams was cut via commercialized heavy duty aluminum frame lettuce cutter (New Star Foodservice, CA, USA), and each cut product is placed immediately in a food container and stored at 4°C with paper towels placed at the top and the bottom of lettuce samples. Before we start to take pictures, the samples will be fetched from the refrigerator one by one. With the container opened, the paper towel on the top will be removed, and the container will be flipped to dump the sample onto the plate. Then, the samples were spread out on the plate via hands with gloves worn to make sure the sample to cover up most of the plate as indicated below in Fig. 1. After taking sample pictures under different illumination conditions, the sample will be put back into the food container and then stored in the refrigerator. Each sample was imaged at 5 different days (0, 1, 3, 5, 8 days after purchasing) to ensure the lettuce showed different browning levels.

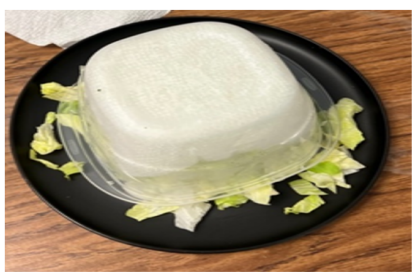
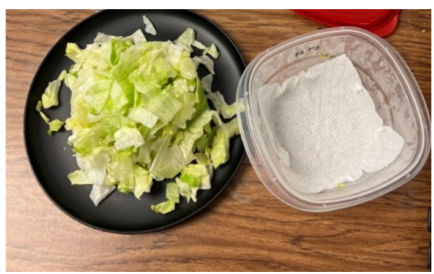
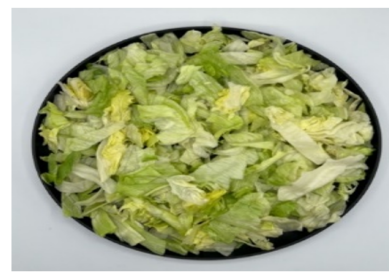

figure 1(a)  
Sample Preparation -Step 1

figure 1(b)  
Sample preparation -Step 2

figure 1(c)  
Sample preparation -Step 3

Figure 1: Sample Preparation

## 1.2 Illumination settings and image acquisition

FVD contains images from 9 lettuce samples with each sample imaged under 15 different illumination conditions considering 5 different illumination temperature (3500K, 4000K, 4250K, 4500K, 4750K) and 3 different illumination power (17.5 W, 20 W, 22.5 W). The lighting condition is generated from a professional photography light box with adjustable brightness and illumination temperature. Each sample was imaged at 5 different days with a total of 675 images (9 samples* 15 images/sample/day * 5 days).

Basler acA1920-40gc camera is mounted on the top of the lighting box to record images with Basler Pylon software, as shown in Fig. 2. Before taking images, the auto white balance and auto exposure are turned off, and related parameters are fixed, with green and red balance ratio as 1, and blue balance ratio as 2.25. All samples will be placed in the same position of the lightbox in order to make sure the sample is fully covered in the camera's field of view.

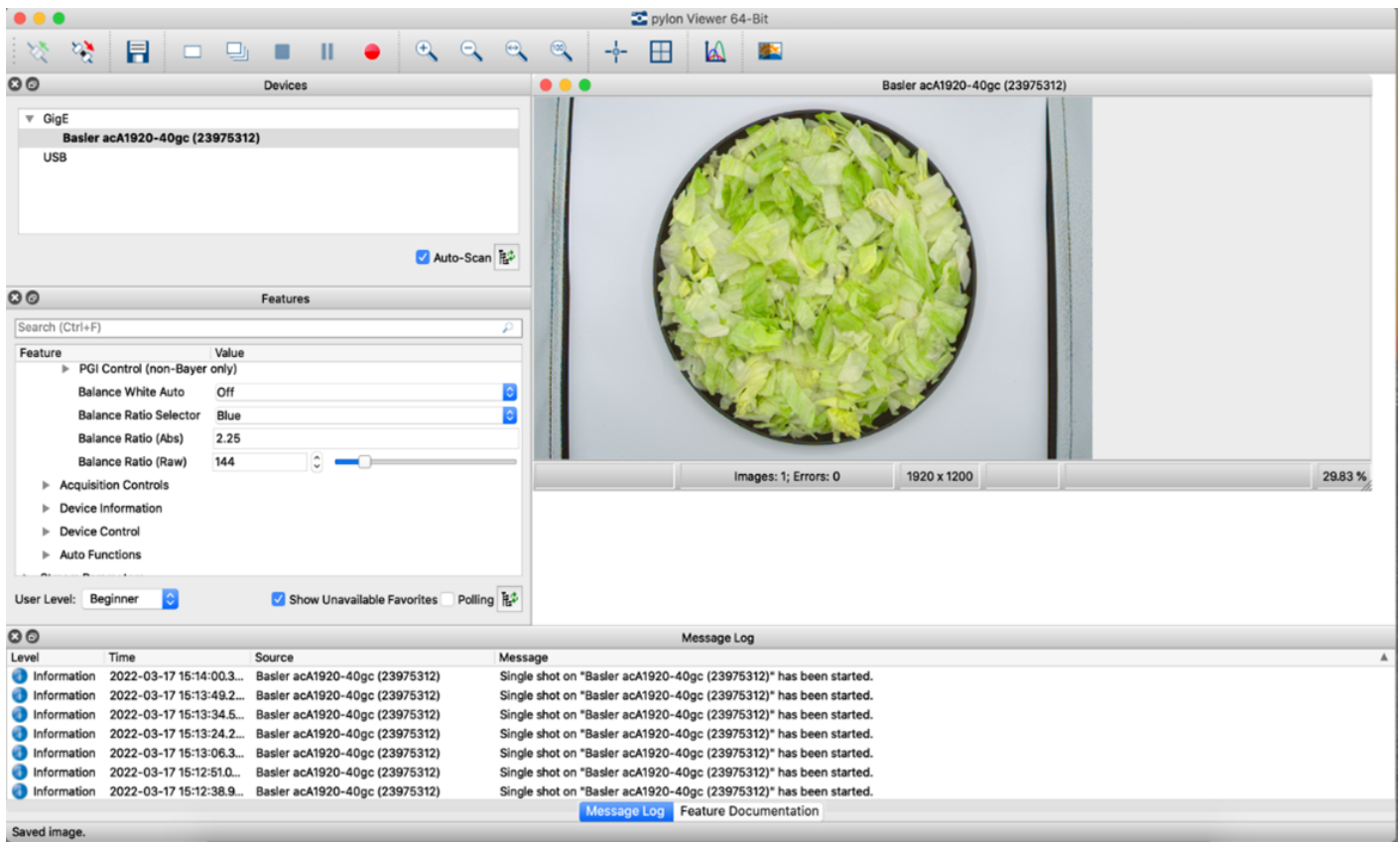

Figure 2: Basler Pylon software interface for image acquisition

For each sample, fifteen images are captured with illumination settings described above, as shown in Table 1. Each image has image size 1024*1024. While saving the image, the image file was named like following example: t5s1day1_3500K_17.5W, where the first field t5 represents that is test 5 (we had conducted several round preliminary tests before the final one), s1 means sample 1, 3500K is the illumination temperature of the image, and 17.5 W is the illumination power setting of the image.

Table 1: Illumination temperature and power settings

| 3500 K, 17.5 W | 4000 K, 17.5 W | 4250 K, 17.5 W | 4500 K, 17.5 W | 4750 K, 17.5 W |
|---|---|---|---|---|
| 3500 K, 20 W | 4000 K, 20 W | 4250 K, 20 W | 4500 K, 20 W | 4750 K, 20 W |

| 3500 K, 22.5 W | 4000 K, 22.5 W | 4250 K, 22.5 W | 4500 K, 22.5 W | 4750 K, 22.5 W |

## 1.3 Data augmentation

To support related deep learning studies, data augmentation can be conducted for the data we collected. As we are working for color constancy and illumination prediction applications, the augmentation operations related to image color values, such gamma-correction, image enhancement, were not adopted. Suggested augmentation operations include rotation, translation, shear, cropping, horizontal/vertical flipping, and mirroring. These operations can be adopted utilizing the Keras (Gulli and Pal, 2017) framework built on top of TensorFlow with the 'ImageDataGenerator' class..

# 2. Results and discussions

## 2.1 Samples images

Some sample images are shown in Fig. 3 and Fig. 4, the effects from illumination power can be observed for various days of freshness from day0 to day 8 of the experiment for constant illumination temperature 4750K, while the effects from illumination temperature can be observed from for a constant power level 17.5 W for various days from day0 to day8.

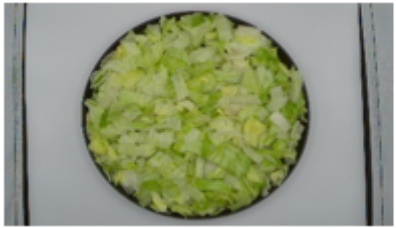
Figure 3(a).
t5s9day0_4750K_17.5W

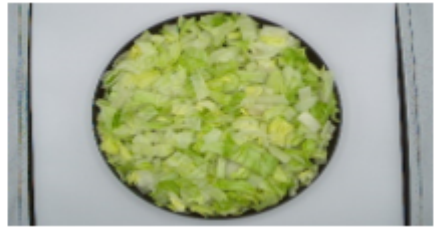
Figure 3(b).
t5s9day0_4750K_20W

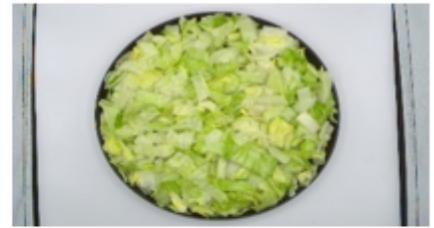
Figure 3(c).
t5s9day0_4750K_22.5W

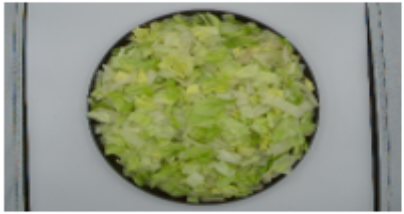
Figure 3(d).
t5s9day1_4750K_17.5W

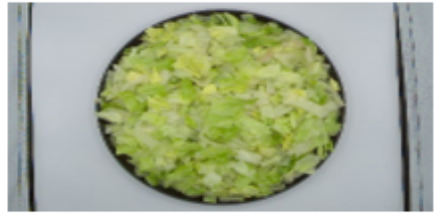
Figure 3(e).
t5s9day1_4750K_20W

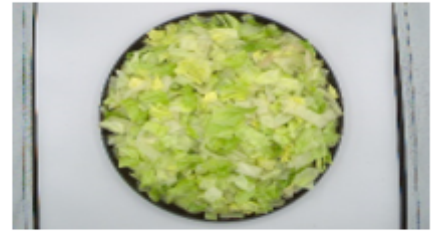
Figure 3(f).
t5s9day1_4750K_22.5W

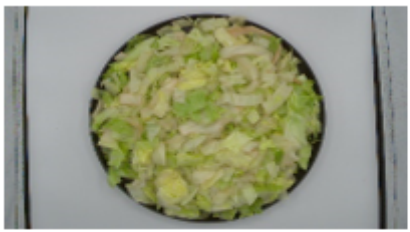
Figure 3(g).
t5s9day3_4750K_17.5W

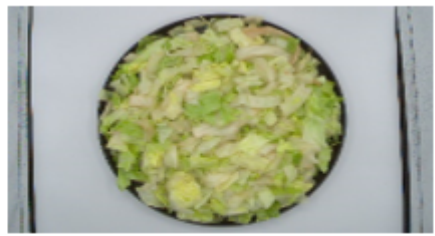
Figure 3(h).
t5s9day3_4750K_20W

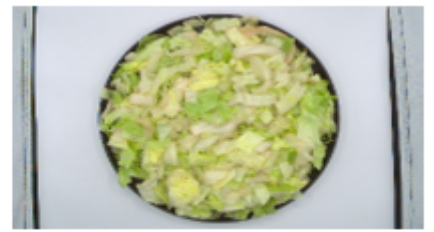
Figure 3(i).
t5s9day3_4750K_22.5W

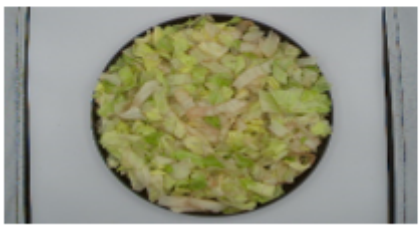
Figure 3(j).
t5s9day5_4750K_17.5W

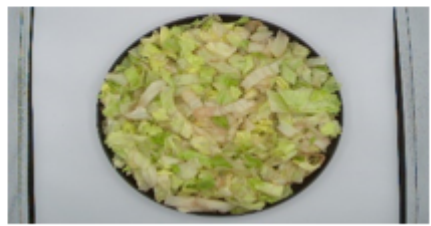
Figure 3(k).
t5s9day5_4750K_20W

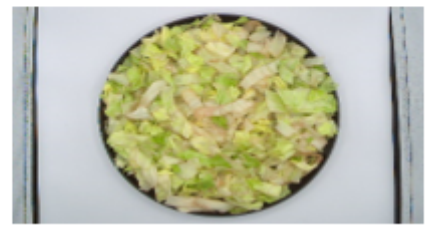
Figure 3(l).
t5s9day5_4750K_22.5W

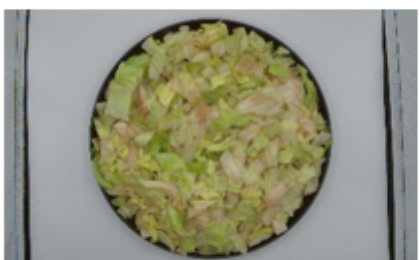
Figure 3(m).
t5s9day8_4750K_17.5W

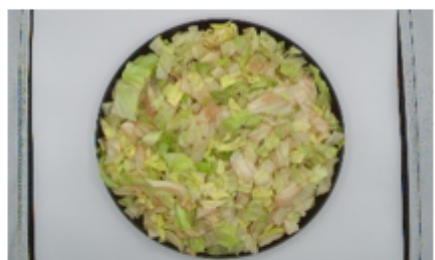
Figure 3(n).
t5s9day8_4750K_20W

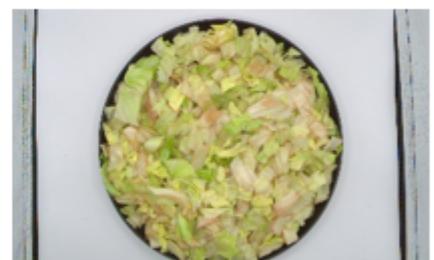
Figure 3(o).
t5s9day8_4750K_22.5W

Figure 3: Variations in illumination power

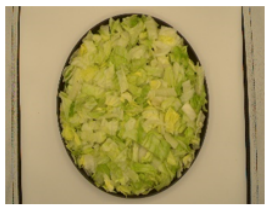
Figure 4 (a)
t5s9day0_3500K_
17.5W

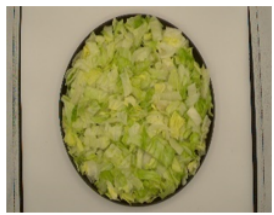
Figure 4 (b)
t5s9day0_4000K_
17.5W

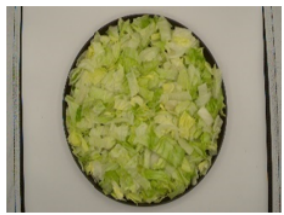
Figure 4 (c)
t5s9day0_4250K_
17.5W

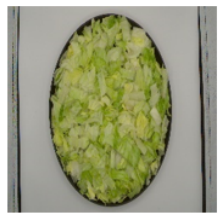
Figure 4 (d)
t5s9day0_4500K_
17.5W

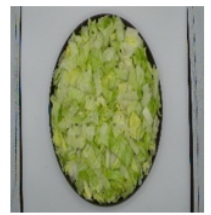
Figure 4 (e)
t5s9day0_4750K_
17.5W

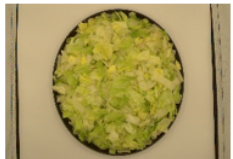
Figure 4 (f)
t5s9day1_3500K_
17.5W

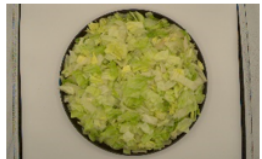
Figure 4 (g)
t5s9day1_4000K_
17.5W

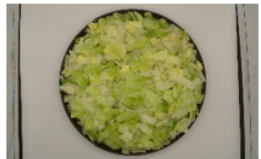
Figure 4 (h)
t5s9day1_4250K_
17.5W

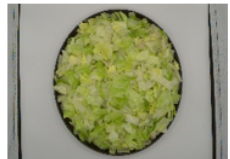
Figure 4 (i)
t5s9day1_4500K_
17.5W

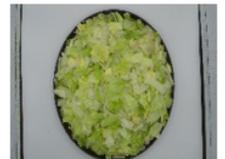
Figure 4 (j)
t5s9day1_4750K_
17.5W

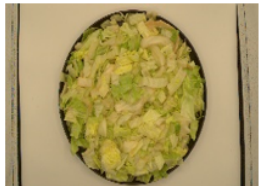
Figure 4 (k)
t5s9day3_3500K_
17.5W

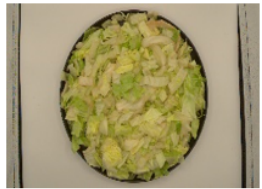
Figure 4 (l)
t5s9day3_4000K_
17.5W

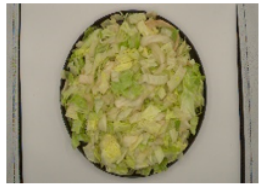
Figure 4 (m)
t5s9day3_4250K_
17.5W

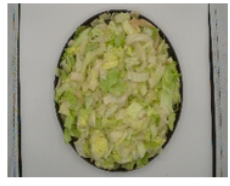
Figure 4 (n)
t5s9day3_4500K_1
7.5W

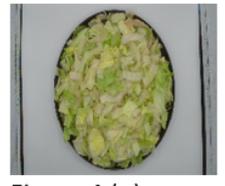
Figure 4 (o)
t5s9day3_4750K_
17.5W

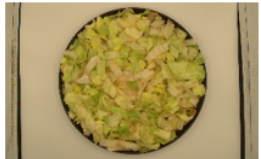
Figure 4 (p)
t5s9day5_3500K_
17.5W

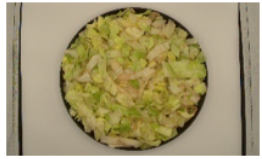
Figure 4 (q)
t5s9day5_4000K_
17.5W

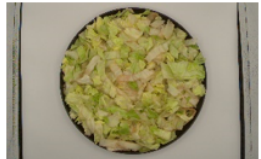
Figure 4 (r)
t5s9day5_4250K_
17.5W

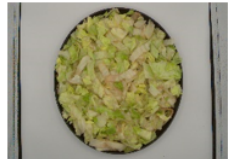
Figure 4 (s)
t5s9day5_4500K_
17.5W

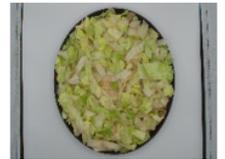
Figure 4 (t)
t5s9day5_4750K_
17.5W

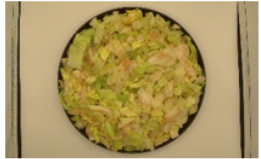
Figure 4 (u)
t5s9day8_3500K_
17.5W

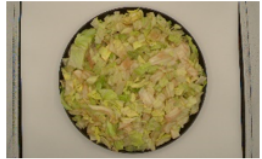
Figure 4 (v)
t5s9day8_4000K_
17.5W

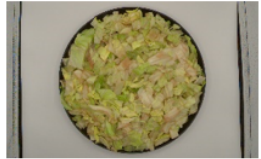
Figure 4 (w)
t5s9day8_4250K_
17.5W

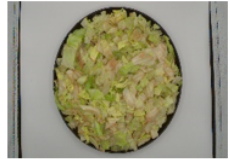
Figure 4 (x)
t5s9day8_4500K_
17.5W

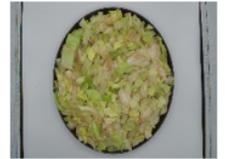
Figure 4 (y)
t5s9day8_4750K_
17.5W

Figure 4: Variations in illumination temperature

## 2.2 Other related datasets

Illumination plays important roles in the computer vision domain. Based on the literature review, we list a series of dataset related to illumination properties estimations (light directions, illuminant type, etc.,), which could benefit researchers working in this area. Our FVD firstly considers the effects from illumination temperature and illumination power on digital images.

(Murmann et al., n.d.)presents a dataset of objects in small and controlled environments for interior scenes captured with a flash from 25 different orientations with constant illumination.

(Xu et al., 2018), similar to (Murmann et al., n.d.), also proposes a dataset of scenes with different light directions. The differences are that the images are rendered and that the light directions are randomized.

(Sun et al., 2019) specifically targets at face relighting, not extending to general scenes, this method is trained on a dataset consisting of 18 individuals captured under different directional light sources under controlled settings, with the individual illuminated by a sphere with numerous lights.

(Helou et al., 2020) aim at developing a Virtual Image Dataset for Illumination Transfer (VIDIT), to create a reference evaluation benchmark and to push forward the development of illumination manipulation methods. VIDIT contains 300 virtual scenes used for training, where every scene is captured 40 times in total: from 8 equally spaced azimuthal angles, each lit with 5 different illuminants (Helou et al., n.d.).

## 2.3 Future research plan

Our dataset can be used in different applications, both for research and applications. From an artificial intelligence perspective, FVD can be used to predict the power of the illuminant in a scene, as well as its color temperature. With the information, a given input scene can be transformed into a variety of illumination settings as needed. In addition, it can be considered to improve the interpretability and robustness of deep learning models. From a sensory science perspective, FVD can serve as a benchmark dataset to describe the relationship between food appearance and consumer acceptances under different illumination conditions. The techniques to remove illumination effects can also be considered. The developed techniques are expected to benefit not only the specific food industry application, but also potential other industrial inspection applications.

## 2.4 Dataset accessibility

This dataset is the first one to describe the relationship among food appearance, consumer acceptances, and external illumination conditions (lighting temperature and power). This dataset can be properly requested from the team for benefiting related academic studies about data science, food science, and marketing studies.

# Acknowledgement

The research is supported by the National Science Foundation under the grant under Award No. OIA-1946391.We would like to thank our graduate & senior undergraduate students Olivia Torres, Robert Blindauer and Yihong Feng for helping us collect, analyze and grade samples.

# References

Andrew-LaVigne-House-Ag-testimony-7-12-17.pdf, n.d.

Gulli, A., Pal, S., 2017. Deep Learning with Keras. Packt Publishing.

Hasenbeck, A., Cho, S., Meullenet, J.F.C., Tokar, T., Yang, F.L., Huddleston, E.A., Seo, H.-S., 2014. Color and illuminance level of lighting can modulate willingness to eat bell peppers. J. Sci. Food Agric. 94 10, 2049–56.

Helou, M.E., Zhou, R., Barthas, J., Süsstrunk, S., 2020. VIDIT: Virtual Image Dataset for Illumination Transfer. CoRR abs/2005.05460.

Helou, M.E., Zhou, R., Yazdani, A., n.d. VIDIT Dataset.

LeCun, Y., Bengio, Y., Hinton, G., 2015. Deep learning. Nature 521, 436–444. https://doi.org/10.1038/nature14539

Murmann, L., Gharbi, M., Aittala, M., Durand, F., n.d. A Dataset of Multi-Illumination Images in the Wild 10.

Murray, R.F., Adams, W.J., 2019. Visual perception and natural illumination. Vis. Percept. 30, 48–54. https://doi.org/10.1016/j.cobeha.2019.06.001

Sun, T., Barron, J.T., Tsai, Y.-T., Xu, Z., Yu, X., Fyffe, G., Rhemann, C., Busch, J., Debevec, P.E., Ramamoorthi, R., 2019. Single Image Portrait Relighting. CoRR abs/1905.00824.

Wang, H., Cuijpers, R.H., Luo, M.R., Heynderickx, I., Zheng, Z., 2015. Optimal illumination for local contrast enhancement based on the human visual system. J. Biomed. Opt. 20, 015005. https://doi.org/10.1117/1.JBO.20.1.015005

Xu, Z., Sunkavalli, K., Hadap, S., Ramamoorthi, R., 2018. Deep image-based relighting from optimal sparse samples. ACM Trans. Graph. TOG 37, 1–13.